\documentclass[manuscript,nonacm]{acmart}
\AtBeginDocument{%
  }

\usepackage{threeparttable}
\usepackage{tcolorbox}
\usepackage{booktabs}
\usepackage{float}
\title[NSV-Shift]{NSV-Shift: A Contrastive Benchmark for
Non-Speech Vocalization Understanding and
Response Adaptation in Speech-to-Speech Models}

\begin{document}


\author{Ziwei Chen}
\email{nic@ucsd.edu}
\affiliation{%
  \institution{University of California San Diego}
  \city{La Jolla}
  \state{California}
  \country{USA}
}

\noindent
{\fontsize{14}{18}\selectfont\bfseries
NSV-Shift: A Contrastive Benchmark for Non-Speech Vocalization Understanding
and Response Adaptation in Speech-to-Speech Models
\par}

\vspace{0.5cm}

\noindent
{\Large Ziwei Chen\footnote{This work was conducted during the Eigen Labs Fellowship.}\par}

\noindent
{\Large University of California San Diego\par}

\vspace{0.7cm}

\noindent
{\bfseries Abstract\par}

\vspace{0.25cm}

{
\setlength{\parindent}{0pt}
\noindent
We introduce NSV-Shift, a contrastive benchmark for evaluating whether
speech-to-speech models can understand non-speech vocalizations (NSVs)
and adapt their responses accordingly. Each pair contains two conversations
with identical lexical content that differ only in the NSV embedded in
the final turn. Our pilot contains 22 human-verified pairs (44 audio
conditions) and evaluates five models on NSV perception, emotion
understanding, and response adaptation. Results show that models generally
perform better at detecting NSVs than at interpreting their fine-grained
emotional meaning or producing appropriately differentiated responses.
The data construction pipeline, dataset, and evaluation pipeline are
publicly available at
\url{https://github.com/ChenzwNina/nsv-construction}.
}


\section{Introduction}
\label{section:overview}

As large language models (LLMs) become increasingly
integrated into our daily lives, AI may eventually
participate in spoken conversations as a personal
assistant or even a companion. To make this possible,
an important question is whether audio LLMs can
understand people's emotions and adapt their responses
accordingly. Speech conveys rich emotional information through prosody, tone, volume, and other vocal cues. One important aspect of spoken communication that is largely absent from text is non-speech vocalizations
(NSVs). NSVs are non-lexical vocal sounds, such as laughter, sighs, and gasps, that can convey emotional
meaning without words. They are also common in
everyday conversations. For example, a study of the
ICSI meeting corpus reported that laughter accounted
for approximately 9\% of vocalization time in
multi-speaker conversations
\cite{laskowski2011predicting}. These observations motivate the need to evaluate whether speech-to-speech (S2S) models can perceive
NSVs and understand the emotions they convey.

Despite their importance, benchmarks for NSV
understanding remain underdeveloped. Previous studies have examined how S2S models perceive changes in prosody and tone, and whether these changes influence
model behavior
\cite{yang2026paras2s, zhou2026echomind,
liu2026hear2act}. Research specifically focused on
NSVs has largely emphasized recognition and
classification, with less attention to how models
interpret NSVs within conversations and adapt their responses accordingly. This leaves an important
research gap. To address this gap, we propose NSV-Shift, a benchmark that systematically evaluates whether S2S models can
\textit{(1) perceive NSVs embedded in spoken
conversations, (2) interpret the emotions they convey in context, and (3) adapt their response content
appropriately.} 

\section{Benchmark Structure}
NSV-Shift is a contrastive benchmark for evaluating how S2S models understand NSVs in conversational context and adapt their responses accordingly. Each benchmark instance consists of a \textbf{counterfactual conversation pair}. The two versions share the same four-turn dialogue between Speaker 1 and Speaker 2. In the final turn, Speaker 2 first produces an NSV and then says the same lexical sentence in both versions. The only difference between the two conversations is the NSV. The lexical content is intentionally designed to be emotionally underdetermined, such that multiple interpretations remain plausible from the words alone. The two NSVs are selected to support meaningfully different emotional interpretations of Speaker 2's final turn. This design isolates the contribution of the NSV while holding the conversational context and lexical content constant.

The current evaluation set contains \textbf{22 contrastive pairs, or 44 audio conversations}. The NSV inventory includes laughter, crying, sighing, yawning, gasp, groan, and screaming. The verified set covers 15 everyday conversation topics.

\section{Data Construction}
\subsection{Transcript Generation}

We construct candidate conversations using GPT-4o. Each conversation contains four turns:

\begin{itemize}
    \item Turn 1: Speaker 1
    \item Turn 2: Speaker 2
    \item Turn 3: Speaker 1
    \item Turn 4: Speaker 2 produces an NSV followed by a lexical utterance
\end{itemize}

For each candidate pair, the two versions must use identical lexical content and differ only in the NSV at the beginning of the final turn. We generate candidates across 20 everyday conversation topics and all pairwise combinations of seven NSV types. This produces $20 \times \binom{7}{2} = 420$ candidate pairs per generation round. We conduct two independent generation rounds using the same generation grid but different sampled conversations.

The generation prompt enforces four main constraints:

\begin{enumerate}
    \item the lexical content must be identical across the two NSV conditions;
    \item no NSV appears in the first three turns;
    \item Speaker 2's final emotion must remain underdetermined from the lexical content alone; and
    \item the two NSVs must support distinct but socially plausible interpretations of the final turn.
\end{enumerate}

\subsection{Automatic Filtering}

We apply a multi-stage automatic filtering pipeline using GPT-4o to remove low-quality or ambiguous candidates like prior works~\cite{yang2026paras2s}.

\begin{enumerate}
    \item \textbf{Lexical neutrality.}
    We remove conversations in which the lexical content already strongly determines Speaker 2's emotion.

    \item \textbf{NSV reasonability.}
    Each NSV condition is evaluated independently to determine whether the vocalization could naturally occur in the given conversational context. Both conditions must pass.

    \item \textbf{Emotion consistency.}
    Each condition is labeled three independent times using the 27 emotion categories from the GoEmotions taxonomy~\cite{demszky-etal-2020-goemotions}. We retain a condition only when the same emotion label is produced in all three runs.

    \item \textbf{Emotion contrast.}
    We further map the fine-grained emotion labels into the sentiment-level
    grouping defined in GoEmotions~\cite{demszky-etal-2020-goemotions}:
    positive, negative, ambiguous, and neutral. For constructing contrastive
    pairs, we exclude \textit{neutral} and \textit{realization}, and retain
    only pairs whose two variants belong to different sentiment groups.
\end{enumerate}

\subsection{Human Verification}

The remaining candidates undergo human verification. One annotator sees one audio condition at a time, with the two members of a pair presented separately and in randomized order. For each condition, the annotator selects the emotion conveyed by Speaker 2's final turn from the 27 GoEmotions labels, together with \textit{Neutral} and \textit{Unclear}. A pair enters the final evaluation set only when the human emotion label for \textbf{both} conditions exactly matches the corresponding emotion label established during automatic filtering. Across two construction rounds, this process produces \textbf{22 verified contrastive pairs (44 audio conditions)}. The NSVs embedded in the 22 pairs and the emotions they convey are listed in Appendix~\ref{app:pilot_composition}.

\subsection{Audio Construction}

To ensure that the two conditions differ only in their NSVs, we reuse the same synthesized lexical audio within each pair. Turns 1--3 and the lexical portion of Turn 4 are synthesized with
ElevenLabs~\cite{elevenlabs2026audiotags} using two fixed speaker voices (one male and one female). NSVs are generated separately: six NSV types are synthesized using
ElevenLabs audio tags, while gasp is generated using Dia-1.6B with voice cloning~\cite{nari2025dia}. We use WhisperX~\cite{bain2023whisperx} forced alignment to identify
the onset of the first lexical word in the generated NSV segment. The NSV is separated from the following lexical speech and then concatenated with the shared Turn 4 utterance. The complete four-turn conversation is assembled with 150-ms gaps between turns. As a result, the two audio conditions within a pair share the same speakers, lexical content, and conversational timing as closely as possible, while differing in the target NSV.

\section{Evaluation Tasks and Gold Labels}
\subsection{Evaluation Tasks}

NSV-Shift evaluates four stages of NSV understanding, ranging from
perception to conversational response adaptation.

\paragraph{Q1: NSV Perception.}
We first test whether the model can identify the NSV present in the
conversation. The model answers a seven-way multiple-choice question
over the benchmark NSV inventory. The gold label is the NSV used during
audio construction. Random-choice accuracy is 14.3\%.

\paragraph{Q2: Last-Turn Polarity.}
We next test whether the model captures the coarse emotional meaning of
Speaker 2's final turn. The model predicts whether the final-turn
emotion is \textit{negative} or \textit{non-negative}. The gold answer
is derived from the verified fine-grained emotion label using the
GoEmotions sentiment grouping~\cite{demszky-etal-2020-goemotions}.
Random-choice accuracy is 50\%.

\paragraph{Q3: Fine-Grained Emotion Recognition.}
We then evaluate whether the model can identify the specific emotion
conveyed by Speaker 2's final turn. The model selects from the 27
GoEmotions emotion categories~\cite{demszky-etal-2020-goemotions},
together with \textit{Neutral} and \textit{Unclear}. The gold answer is
the locked emotion label produced during benchmark construction and
subsequently confirmed through human verification. Random-choice
accuracy is approximately 3.4\%.

\paragraph{Q4: Response Adaptation.}
Finally, we test whether changing only the NSV leads the model to
produce a correspondingly different and contextually appropriate
response. For each counterfactual pair, the model generates a spoken
next-turn response as Speaker 1 for both NSV conditions. Rather than comparing each response against a single reference answer,
we evaluate the two responses jointly using counterfactual matching.
A judge receives the two conversational contexts and the two generated
responses, with both context and response order randomized, and
determines the best one-to-one assignment between them. The model
succeeds when the judge recovers the original context--response pairing.
The judge may also return a tie. Without ties, random assignment
corresponds to a 50\% baseline.

\subsection{Scoring Protocol}

For Q1--Q3, we extract the model's spoken multiple-choice answer and
score it using exact match against the corresponding gold label. For Q4, we use \texttt{gpt-5.6-terra} as the matching judge. The judge
returns one of three outcomes: correct pairing, swapped pairing, or tie.
Ties are counted as incorrect in the reported accuracy. GPT-4o-generated Speaker 1 replies are retained as qualitative
references, but they are not used as gold answers for Q4. This avoids
rewarding similarity to one particular response and instead measures
whether the model's own responses reflect the difference introduced by
the NSV.

\subsection{Evaluation Models}
We evaluate NSV-Shift on five speech-to-speech models:

\begin{itemize}
    \item \textbf{gpt-realtime-2.1} (OpenAI Realtime)
    \item \textbf{gemini-3.8-live} (Gemini Live)
    \item \textbf{gpt-live} (\texttt{gpt-live-1})
    \item \textbf{qwen-audio-3.0} (\texttt{qwen-audio-3.0-realtime-plus})
    \item \textbf{grok-voice-think-fast-2.0} (xAI Realtime)
\end{itemize}

Each model is evaluated on all 44 audio conditions for Q1--Q3 and all
22 counterfactual pairs for Q4. Each question is tested in a separate
session to avoid information from one task influencing another.

\section{Evaluation Results}
All five models were evaluated on the complete pilot set. Table~\ref{tab:main_results}
reports overall performance on the 44 audio conditions for Q1--Q3 and
the 22 counterfactual pairs for Q4.

\subsection{Overall Performance}

\begin{table}[t]
\centering
\small
\begin{tabular}{lcccccc}
\toprule
\textbf{Model} &
\textbf{Q1} &
\textbf{Q2} &
\textbf{Q3} &
\textbf{Q4} &\\
\midrule

gpt-realtime-2.1
& \textbf{38/44 (86\%)}
& 35/44 (80\%)
& 17/44 (39\%)
& \textbf{14/22 (64\%)}\\

gemini-3.8-live
& 33/44 (75\%)
& 35/44 (80\%)
& \textbf{26/44 (59\%)}
& 13/22 (59\%)\\

gpt-live
& 32/44 (73\%)
& \textbf{37/44 (84\%)}
& 13/44 (30\%)
& 11/22 (50\%)\\

qwen-audio-3.0
& 36/44 (82\%)
& 32/44 (73\%)
& 14/44 (32\%)
& 7/22 (32\%)\\

grok-voice-think-fast-2.0
& 12/44 (27\%)
& 26/44 (59\%)
& 7/44 (16\%)
& 7/22 (32\%)\\

\bottomrule
\end{tabular}
\caption{Overall performance on NSV-Shift. Q1--Q3 are evaluated over
44 audio conditions, while Q4 is evaluated over 22 counterfactual pairs.}
\label{tab:main_results}
\end{table}

Performance decreases as the task moves from coarse perception toward
more context-sensitive interpretation and response adaptation.
gpt-realtime achieves the highest Q1 accuracy (86\%) and Q4 matching
accuracy (64\%), while Gemini achieves the highest fine-grained emotion
accuracy on Q3 (59\%). gpt-live performs best on coarse polarity
classification (84\%) but reaches 50\% on Q4. Qwen performs strongly on
NSV perception (82\%) but obtains 32\% on response matching. Grok shows
the lowest Q1 and Q3 accuracy, at 27\% and 16\%, respectively.

\subsection{NSV Perception}

Q1 performance varies substantially across vocalization types.
Table~\ref{tab:nsv_perception} pools predictions across the five models.

\begin{table}[t]
\centering
\small
\begin{tabular}{lrrrrrrrr}
\toprule
\textbf{NSV} &
\textbf{n} &
\textbf{Correct} &
\textbf{Rate} &
\textbf{Realtime} &
\textbf{Gemini} &
\textbf{GPT-Live} &
\textbf{Qwen} &
\textbf{Grok} \\
\midrule

sighing   & 35 & 28 & 80\% & 6/7   & 7/7   & 7/7   & 6/7   & 2/7 \\
laughter  & 85 & 66 & 78\% & 16/17 & 14/17 & 12/17 & 15/17 & 9/17 \\
gasp      & 60 & 46 & 77\% & 12/12 & 11/12 & 10/12 & 12/12 & 1/12 \\
crying    & 5  & 3  & 60\% & 1/1   & 1/1   & 0/1   & 1/1   & 0/1 \\
screaming & 10 & 3  & 30\% & 2/2   & 0/2   & 1/2   & 0/2   & 0/2 \\
groan     & 25 & 5  & 20\% & 1/5   & 0/5   & 2/5   & 2/5   & 0/5 \\

\bottomrule
\end{tabular}
\caption{Q1 perception performance by NSV type, pooled across five models.}
\label{tab:nsv_perception}
\end{table}

Laughter, gasp, and sighing are recognized relatively consistently by
the four non-Grok models. Groan is substantially more difficult, with
only 5 of 25 pooled predictions correct. Screaming also has low observed
accuracy (3/10), although only two screaming clips are present in the
evaluation set. Grok is notably different from the other models on gasp:
it identifies only 1 of 12 gasp clips correctly, compared with 10--12
correct predictions for each of the other four models.

\subsection{Emotion Interpretation}

The gap between coarse and fine-grained emotion understanding is visible
in the overall results: Q2 accuracy ranges from 59\% to 84\%, whereas
Q3 ranges from 16\% to 59\%. Table~\ref{tab:emotion_results} further breaks down Q3 performance by the verified emotion label.

\begin{table}[t]
\centering
\small
\begin{tabular}{lrrrl}
\toprule
\textbf{Emotion} &
\textbf{n} &
\textbf{Correct} &
\textbf{Rate} &
\textbf{Typical NSV} \\
\midrule

surprise       & 60 & 41 & 68\% & gasp \\
disappointment & 30 & 13 & 43\% & sighing \\
amusement      & 85 & 18 & 21\% & laughter \\
annoyance      & 30 & 4  & 13\% & groan \\
excitement     & 10 & 1  & 10\% & screaming \\
sadness        & 5  & 0  & 0\%  & crying \\

\bottomrule
\end{tabular}
\caption{Q3 fine-grained emotion accuracy pooled across five models.}
\label{tab:emotion_results}
\end{table}

Surprise is recognized most consistently, with 41 of 60 pooled predictions correct. In contrast, amusement is selected correctly in
only 18 of 85 cases despite laughter being correctly perceived in
66 of 85 corresponding Q1 predictions. Models choose surprise rather than the correct amusement in 26 our of 85 cases. Annoyance is also difficult and coincides with poor groan perception.
Results for sadness and excitement should be interpreted cautiously
because the benchmark contains very few crying and screaming examples.

\subsection{Response Adaptation}

Q4 evaluates whether the NSV affects the model's subsequent conversational
behavior. gpt-realtime obtains the highest matching accuracy among the
audio models at 64\%, followed by Gemini at 59\%. gpt-live obtains 50\%,
while Qwen and Grok each obtain 32\%. The distribution of swapped responses
provides additional evidence that models often produce responses that are
insufficiently differentiated between the two NSV conditions. Qwen produces
13 swapped pairs compared with 7 correct matches, while Grok produces
12 swaps, 7 correct matches, and 3 ties.

To provide a reference for response generation when the conversational
content and NSV information are represented textually, we additionally
evaluate two text-based LLM writers on the same 22 counterfactual pairs.
Unlike the speech-to-speech models, these models do not need to infer the
NSV directly from audio; instead, they generate responses from the textual
representation of each condition. We then apply the same Q4 matching
evaluation to determine whether their two generated responses are correctly
aligned with the corresponding counterfactual conditions. As shown in Table~\ref{tab:text_writer_q4}, both text-based writers achieve
higher Q4 matching accuracy than the strongest audio model: GPT-4o reaches
68\%, while Claude Opus 5 reaches 77\%, compared with 64\% for gpt-realtime. One possible explanation is that stronger general-purpose language models are better able to generate sufficiently differentiated responses for contrasting emotional conditions. The lower performance of the audio language models may therefore reflect not only difficulty in perceiving and interpreting paralinguistic cues, but also limitations in their response-generation capabilities relative to stronger text-based models.

\begin{table}[t]
\centering
\small
\begin{tabular}{lc}
\toprule
\textbf{Text Writer} &
\textbf{Q4} \\
\midrule

GPT-4o
& 15/22 (68\%) \\

Claude Opus 5
& \textbf{17/22 (77\%)} \\

\bottomrule
\end{tabular}
\caption{Q4 response-matching performance of text-based response writers
on the 22 counterfactual pairs.}
\label{tab:text_writer_q4}
\end{table}

Overall, the Q4 results suggest that perceiving and interpreting an NSV does not guarantee that the signal will meaningfully influence the model's next-turn response. The stronger
performance of the text-based writers further suggests that part of the
difficulty for speech-to-speech models may arise before or during the
translation of paralinguistic information into response behavior.

\paragraph{Summary.}
Across the four tasks, we observe a gap between \textit{perceiving} an
NSV, \textit{interpreting} its emotional meaning, and \textit{using} it
to adapt a conversational response. Several models can reliably recognize
common NSVs and their coarse emotional polarity, while fine-grained
emotion recognition and counterfactual response adaptation remain more
challenging.

\section{Limitations}
The current pilot also contains only eight distinct NSV--emotion contrast
types, with most pairs involving laughter. We therefore do not yet know
whether the observed differences between models generalize to a broader
space of vocalization contrasts. Expanding both the number of NSVs and
the diversity of within-NSV meanings is an important direction for scaling
NSV-Shift beyond the present pilot.

\clearpage

\bibliographystyle{ACM-Reference-Format}
\bibliography{main}

\appendix
\section{Verified Audio Pairs Composition}
The final pilot set contains 22 human-verified counterfactual pairs,
corresponding to 44 audio conditions. Each pair shares the same lexical
conversation but differs in the non-speech vocalization (NSV) inserted
into the target turn. Table~\ref{tab:verified_pairs} lists the full set
of verified pairs and their locked emotion labels.

\begin{table*}[t]
\centering
\small
\begin{tabular}{rllll}
\toprule
\textbf{ID} &
\textbf{Topic} &
\textbf{NSV 1} &
\textbf{NSV 2} &
\textbf{Locked Emotions} \\
\midrule

1   & Childhood Memories        & crying   & gasp      & sadness / surprise \\
10  & Childhood Memories        & laughter & gasp      & amusement / surprise \\
31  & Cultural Differences      & laughter & gasp      & amusement / surprise \\
73  & Dreams \& Sleep           & laughter & gasp      & amusement / surprise \\
116 & Entertainment             & laughter & groan     & amusement / annoyance \\
178 & Food \& Drinks            & laughter & gasp      & amusement / surprise \\
223 & Health \& Fitness         & laughter & sighing   & amusement / disappointment \\
286 & Internet Culture          & laughter & sighing   & amusement / annoyance \\
322 & Personal Life             & gasp     & screaming & surprise / excitement \\
347 & Relationships             & laughter & groan     & amusement / annoyance \\
367 & Technology \& Gadgets     & laughter & gasp      & amusement / surprise \\
386 & Travel                    & groan    & screaming & annoyance / excitement \\
414 & Work \& Studies           & sighing  & gasp      & disappointment / surprise \\
\midrule
110 & Internet Culture          & laughter & groan     & amusement / annoyance \\
131 & Travel                    & laughter & groan     & amusement / annoyance \\
149 & Food \& Drinks            & laughter & sighing   & amusement / disappointment \\
254 & Finance \& Money          & laughter & sighing   & amusement / disappointment \\
275 & Fashion \& Style          & laughter & sighing   & amusement / disappointment \\
277 & Fashion \& Style          & laughter & gasp      & amusement / surprise \\
319 & Cultural Differences      & laughter & gasp      & amusement / surprise \\
361 & Holidays \& Celebrations  & laughter & gasp      & amusement / surprise \\
370 & Holidays \& Celebrations  & sighing  & gasp      & disappointment / surprise \\

\bottomrule
\end{tabular}
\caption{The 22 counterfactual pairs in the verified pilot set. Each pair
contains two acoustically distinct NSV conditions while holding the lexical
conversation fixed. The first 13 rows were obtained in the first
construction round and the remaining 9 in the second round.}
\label{tab:verified_pairs}
\end{table*}
\label{app:pilot_composition}

\end{document}